\documentclass[]{article}
\usepackage{proceed2e}

\usepackage{times}
\usepackage{graphicx} % more modern
\usepackage{subfigure} 
\usepackage{amsmath}
\usepackage{amsfonts}
\usepackage{amssymb}
\usepackage{pgfplotstable}
\usepackage{pgfplots}
\usepackage{mathtools}

\usepackage{natbib}

\usepackage{algorithm}
\usepackage{algorithmic}
\usepackage{lmodern}

\usepackage{color}

\usepackage{hyperref}

\title{Transfer from Multiple Linear \\ 
           Predictive State Representations (PSR)}

\author{} % LEAVE BLANK FOR ORIGINAL SUBMISSION.
\author{ {\bf Sri Ramana Sekharan\thanks{$^{*}$Robotics Department. Worcester Polytechnic Institute. Worcester, MA 01609}} \\
ssekharan@wpi.edu\\
\And
{\bf Ramkumar Natarajan$^{*}$}  \\
rnatarajan@wpi.edu\\
\And
{\bf Siddharthan Rajasekaran$^{*}$}   \\
sperundurairajas@wpi.edu\\
}

\begin{document}

\maketitle
%%%%%%%%%%%%%%%%%%%%%%%%%%%%%%%%%%%%%%%%%%
\begin{abstract}
\textbf{In this paper we tackle the problem of transferring policy from multiple partially observable source environments to a partially observable target environment modeled as predictive state representation. This is an entirely new approach with no previous work, other than the case of transfer in fully observable domains. We develop algorithms to successfully achieve policy transfer when we have the model of both the source and target tasks and discuss in detail their performance and shortcomings. These algorithms could be a starting point for the field of transfer learning in partial observability. } 
\end{abstract}

%%%%%%%%%%%%%%%%%%%%%%%%%%%%%%%%%%%%%%%%%%%%%
\section{INTRODUCTION}
The goal of reinforcement learning is to come up with a way to optimally interact with the environment, modeled as a sequential decision making process. In case of fully observable domains, the environment is modelled as a Markov decision Process. Off the shelf, algorithms like Q-learning are designed to find the optimal policy in such domains. In the case of partially observable domains, the environment is modeled as a Partially Observable Markov Decision Process (POMDPs). Solving a POMDP is still a field of research due to the non-scalability of value iteration like approaches. One of the big disadvantages of POMDPs is the inherent need to specify the underlying states beforehand. This is because the state representation in POMDPs consist of beliefs over latent set of states. This causes problems since in some sense "states" are an imaginary entities unobservable to the agent.

To alleviate the above problem a new way of modeling partially observable domains known as Predictive State Representation was introduced. In this model, there is no concept of states. The agent using this model learns and plans entirely using only observations. This has a huge advantage over POMDPs in that, not only we do not need to mention the number of states beforehand, but also some systems which cannot be modeled as a POMDP can be modeled as a PSR [8] [10]. This is the main reason we choose Predictive State Representation as our basic framework. 

Recent trends in Reinforcement learning indicate the development of algorithms that can achieve transfer of knowledge across domains. Transfer learning is an active area of research in reinforcement learning. The goal behind transfer learning is to learn how to act in some (possibly smaller) source tasks and transfer that knowledge to perform better in a (possibly larger) target domain. Transfer learning in partially observable domains has not been attempted before. Such an algorithm would find applications in many real-world problems since many of them are partially observable.  

Consider a robot trained to grasp a series of regular objects like a sphere, a pyramid and a cube with noisy sensors. Our aim is to model each of these tasks as a PSR. With the model, the robot now has learned a policy for each of the tasks. Given an object with a very complicated shape, the robot has to learn to grasp the new object from scratch which would take more time and be very inefficient. The motivation behind this work is to develop a framework that uses the already learned regular object grasping policies as input and transfer that knowledge to come up with a policy to grasp the complex object. The advantage of using PSR is their ability to handle uncertainty robustly, so even in the presence of noisy perception system our framework would be able to transfer that knowledge to a larger task. 

In this paper we give a basic formulation for the transferring policy between a set of source tasks and a target task. We model them as Predictive State Representation so that our algorithm applies to both fully and partially observable domains. From the standpoint of the basic formulation we develop two algorithms for transfer, test them in a simple domain and discuss the results and shortcomings in detail in this work. 

In the next section we give a high level overview of related approaches, followed by a mathematical overview of PSRs. Then we formulate the problem mathematically. We then explain our procedure in detail, i.e. how we learn the model, how we plan using the learned model and then explain our approach for transfer. We then show the results of our experiments followed by inferences and directions for future work.
%%%%%%%%%%%%%%%%%%%%%%%%%%%%%%%%%%%%%%%%%%%%%%%%
\section{RELATED WORK}
There are a variety of different approaches available in the transfer learning literature for different environment settings. But all of them address only fully observable domains. [8] presents a good survey of the available transfer learning literature. On the other hand, many real-world problems are partially observable. To our knowledge, none of the previous works address transfer learning between partially observable domains. 

One of the approaches that is most relevant to our goal is transfer learning using bisimulation metrics [3]. The bisimulation metric specifies the extent of similarity between two states in a Markov Decision Process. This metric bounds the difference of value functions between the two states. But no such metrics have been developed for partially observable domains so far. The closest approach available in literature is the development of bisimulation equivalence, which states whether two belief states in a Partially Observable Markov Decision Process are similar [3]. It does not give information about the extent of similarity.    

Various algorithms have been proposed for learning the PSR model. [4] formulated the first discovery algorithm for linear PSRs with reset. Spectral learning provides a framework for both learning and planning simultaneously. This method was exploited by [6] [2] to integrate learning and planning in predictive representations. [4] have also made progress in the problem of planning using a learned model in which they have suggested value iteration and approximate Q-Learning methods. In our approach we use approximate Q-Learning using Cerebellar Model Articulation Controller (CMAC) [1] [5].

\section{PROBLEM STATEMENT}
\subsection{PREDICTIVE STATE REPRESENTATION}

A controlled dynamical partially observable discrete system with a finite set of actions $\mathcal{A}$ can emit a finite number of observations $\mathcal{O}$. An agent in the system executes an action $a_t \in \mathcal{A}$ and perceives an observation $o_t \in \mathcal{O}$. A history $h$ at any point of time in the system is defined as an action-observation sequence that has occurred till that point of time $h=a_1o_1a_2o_2....a_to_t$. A test is defined as an action-observation sequence that may occur in the future $t=a_{t+1}o_{t+1}a_{t+2}o_{t+2}...a_{t+k}o_{t+k}$. The prediction for test $t$ given by $P(t|h)$ is defined as the probability that the test $t$ will occur in the future given the agent's current history is $h$.

A Predictive State Representation is a tuple $\langle \mathcal{O},\mathcal{A},Q,M_{ao},m_{ao},P(Q|\emptyset) \rangle$ where, $\mathcal{O}=\{o_i\}$ is the set of observations, $ \mathcal{A}=\{a_i\}$ is the set of possible actions, $Q=\{q_i\}$ is the set of core tests, $M_{ao}=[m_{aoq_i}]$ is the matrix whose columns are vectors associated with one-step extended core test, $m_{ao}$ is the matrix whose columns are vectors associated with one-step tests, $P(Q|\emptyset)$ is the initial prediction vector. We use the term "state vector", "belief" and "prediction vector" interchangeably to refer $P(Q|h)$. When a new action-observation pair is observed we update the state vector as follows.
\begin{gather}
P(Q|hao)=\frac{P(aoQ|h)}{P(ao|h)}=\frac{M_{ao}P(Q|h)}{m_{ao}P(Q|h)}
\end{gather}

Any $m_t$ for any $k$-length test $t=a_{t+1}o_{t+1}a_{t+2}o_{t+2}...a_{t+k}o_{t+k}$, can be calculated from the PSR model using the expression[10], 

\begin{equation}
m_t=M_{a_{t+1}o_{t+1}}M_{a_{t+2}o_{t+2}}...m_{a_{t+k}o_{t+k}}
\label{mo}
\end{equation}

To do planning in PSR, we incorporate a discrete set of rewards $R=\{r_i\}$ along with the observation [11].
\subsection{INITIAL PROBLEM FORMULATION}

We have a set of source tasks $S =\{S_i\}$ and a target task $T$. These tasks are modeled as a Predictive State Representation, defined by the tuples $\langle \mathcal{O},\mathcal{A},Q^{S_i},M^{S_i}_{ao},m^{S_i}_{ao},R^{S_i} \rangle$ and  $\langle \mathcal{O},\mathcal{A},Q^{T},M^{T}_{ao},m^{T}_{ao},R^{T} \rangle$ respectively. We assume both the source and target tasks has the same discrete observation and action set, and we have access to the full model of either tasks. Action selection is done in the source task $S$ by using $\mathcal{Q}$ function as follows $$\pi^{*S}(b^{S})=\arg\max_{a} \mathcal{Q}^{*}(b^{S},a)$$ Our goal in transfer learning is to transfer the policy from source to the target. More formally, for any state in target $b^{T}$ we have to find an appropriate state in source such that following the optimal policy w.r.t the source would give optimal action for the current state in the target.  Mathematically, we want to find $b^{*S|T}$, such that, $$\pi^{*T}(b^{T})=\pi^{*S}(b^{*S|T})$$ We denote the extent of similarity of a source task $s$ to the target task as a function of $b^{T}$ as $\Upsilon_s(b^{T})$. We sometimes overload the similarity operator with two arguments, $\Upsilon_s(b^s,b^T)$ which represents the extent of similarity of target task in the current state $b^T$ with the source task when its state is $b^s$. 
Since we have a set of source tasks, we have to find the most similar source $S_i$ where $$i=\arg\max_j \Upsilon_j(b^{T})$$ Our goal is to devise an algorithm to find $b^{*S|T}$ and $\Upsilon_i(b^{T})$. The $'*'$ in $b^{*S|T}$ denotes the optimal belief mapping. Throughout the paper we denote the task to which a symbol belongs using superscripts. For example we denote core test $Q$ of task $S$ as $Q^S$.

%%%%%%%%%%%%%%%%%%%%%%%%%%%%%%%%%%%%%%%%%%%%%%%%%

\section{OUR APPROACH}

We first learn the model of the source tasks and target task. Then we find the optimal $Q$ function for each of the source tasks using approximate Q-learning. We then develop and test two of our approaches for transfer on the target task. 

\subsection{LEARNING THE MODEL}

We use Analytical Discovery Learning (ADL) algorithm [4] to build the model of the PSR. Since we have access to the POMDP model of the environment, we can calculate any entry of the $D$ matrix. Initially, we enumerate all entries corresponding to one step tests and histories. We find the one step core tests and histories, by taking the linearly independent columns and rows. This is followed by calculating the entries associated with one step extensions to core tests and histories. This procedure is repeated till the rank of core matrix on two successive iterations remains constant. Then the linearly independent rows and columns are taken as the final core test/histories.

\begin{algorithm}
   \caption{Analytical Discovery Learning (ADL).}
   \label{alg:adl}
\begin{algorithmic}
 \STATE {CoreHistory, $\mathcal{H}$ $\gets$ \{\}}
 \STATE CoreTest, $Q$ $\gets$ \{\}
 \WHILE {$Rank(\mathcal{D}_t) > Rank(\mathcal{D}_{t-1})$}

\FOR{$ao$ in $\mathcal{A}$ x $\mathcal{O}$}
\STATE P($Q\mid \mathcal{H}ao$) $\gets$ \textbf{belUpdate}($b$($\mathcal{H}ao$),$Q$)
\STATE P($aoQ \mid \mathcal{H}$) $\gets$ \textbf{belUpdate}(b($\mathcal{H}$),($aoQ$))
\STATE P($aoQ \mid \mathcal{H}ao$) $\gets$ \textbf{belUpdate}($b$( $ao\mathcal{H}$),($aoQ$))
\STATE \footnotesize{$\mathcal{D}_t$ $\gets$ \textbf{build}$\mathcal{D}$(P($Q\mid \mathcal{H}ao$),P($aoQ \mid \mathcal{H}$),P(ao$Q \mid \mathcal{H}$))}
\STATE $\mathcal{H}$ $\gets$ \textbf{independentRows}($\mathcal{D}_t$)
\STATE $Q$ $\gets$ \textbf{independentColums}($\mathcal{D}_t$)
\ENDFOR
\ENDWHILE
\STATE \textbf{Return} $\mathcal{D}_t$,$Q$,$\mathcal{H}$
\end{algorithmic}
\end{algorithm}

The $m_t$ associated with any test $t$ can be found using the formula, 
$$m_t=P(Q|\mathcal{H})^{-1}P(t|\mathcal{H})$$

where $P(Q|\mathcal{H})$ is the core matrix obtained at the end of the ADL algorithm and $P(t|\mathcal{H})$ is the probability of seeing test $t$ from each of the $\mathcal{H}$ core histories. Using this formula, the model parameters $(M_{aoq},m_{ao})$ of the PSR can be obtained.

\subsection{ONLINE LEARNING USING APPROXIMATE Q-LEARNING}

As the PSR state containing the prediction vectors is continuous and high-dimensional planning has to be done in a continuous space. James \emph{et al.} [5] used Cerebellar Model Articulation Controller (CMAC) [1] as a function approximator to implement $\mathcal{Q}$-learning [9]. This is the planning stage in which the $\mathcal{Q}$ values of the state and action space are approximated in an online fashion. CMAC is a class of sparse coded memory that has $r$ overlapped and offset tilings with each of them having number of edges equal to the length of query variables. Each edge of the tiling spans the entire space of corresponding state variable and is quantized into various levels based on its sparsity and length of the prediction vector. Any combination of state vector components and the action value activates exactly one portion of each tiling known as a tile. Every such tile of the entire network is initialized with a random $\mathcal{Q}$ value that is reinforced over time to learn the optimal value. The $\mathcal{Q}$ value of state $S$ containing the prediction vector and action $a$ is given by,

$$ \mathcal{Q}(S,a) = \sum_t f(n_t (s,a))$$

where $n_t(s,a)$ returns the indices of the tiles that gets activated upon querying and $f(u)$ returns the corresponding value which are summed up to give the state action value. The online $\mathcal{Q}$-learning works by updating the value of appropriate tiles by $\alpha \delta$, where $\alpha$ is the per grid learning rate and $\delta$ is given as,

$$ \delta = \mathcal{R}_{s,a} + \gamma \max_{a'} \mathcal{Q}(s',a') - \mathcal{Q}(s,a)$$

The generalization and resolution of CMAC depends on number of tilings and number of tiles within each tiling. \\

\textbf{Remark:} The sparsity of CMAC grows exponentially with the length of the prediction vector. A domain with $\mathbb{N}, \mathbb{L}$ quantization levels for each state and $\mathbb{T}$ tilings will have $O(\mathbb{TN}^{\mathbb{L}})$ tiles to pick $\mathbb{L}$ features for every iteration. Moreover there is a trade-off between weighting the quantization levels of each dimension and computational efficiency in selecting unequally sized tiles within each tiling. 

\subsection{TRANSFER LEARNING IN PREDICTIVE STATE REPRESENTATION}

\textit{Definition :} We define the projection of a test $t$ onto a task $K$ after history $h$ as $\chi^{K}(t,h)$, the probability of occurrence of $t$ on task $K$ after history $h$
$$\chi^{K}(t,h)=P^{K}(t|h)=m_tP^{K}(Q^K|h)$$
If we have a set of tests $T=\{t_i\}$, with a slight abuse of notation, we represent the projection vector as $$\chi^{K}(T,h)=\{\chi^{K}(t_i,h)\}$$
We interchangeably use the terms "projection of test onto a task" and "projection of a task onto a test". They both mean the same.  

\subsubsection{Core Test Projection Algorithm}

We define $b_p^{S|T}$ as the projection of source core test on target. It is the probability of occurrence of source core test on the target at any point of time. Mathematically,
$$b_p^{S|T}(h)=\chi^{T}(Q^S,h)={\chi^{T}(q_i^{S},h)}$$ We can calculate this by the following expression
$$b_p^{S|T}=P^{T}(Q^{S}|h)=M^{T}_{Q_S}b^{T}$$
where $M^{T}_{Q_s}$ is the matrix that relates the probability of occurrence of source core tests $Q_S$ in target task $T$, whose columns can be calculated using Eq.\ref{mo}. We use this projection matrix $M^{T}_{Q_s}$ for every iteration to select an action. We propose that the magnitude of similarity is proportional to the probability of occurrence of core-test. So we calculate similarity as $\Upsilon_i(b^T)=\sum b_p^{S_i|T}$.
This results in the Core-Test Projection algorithm. 

\begin{algorithm}
   \caption{Core Test Projection Algorithm}
   \label{proj}
\begin{algorithmic}
   \STATE {\bfseries Input:} \\ Source set $S= \{S_i=\langle M^{S_i},Q^{S_i},P(Q^{S_i}|\emptyset) \rangle \}$ \\ Target $T=\langle M^T,Q^T,P(Q^T|\emptyset) \rangle$
   \FOR{$S_i$ in $S$}
   \STATE $M^{T}_{Q^{S_i}} \gets \textbf{calculateModelMatrix}(Q^{S_i},T)$ \COMMENT{\ref{mo}}
   \ENDFOR
   \STATE $b^{T} \gets P(Q^T|\emptyset)$
   \REPEAT
   \FOR{$S_i$ in $S$}
   \STATE $b_p^{S_i|T} \gets M^{T}_{Q^{S_i}}b^{T}$
   \STATE $\Upsilon_i \gets \sum b_p^{S_i|T}$
   \ENDFOR
   \STATE $BestS \gets \arg\max_i(\Upsilon_i)$
   \STATE $a_t \gets \arg\max_a Q^{*BestS}(b_p^{BestS|T},a)$
   \STATE $\textbf{takeaction}(a_t)$
   \STATE $o_t \gets \textbf{ReceiveObservation}(a_t)$
   \STATE $b^T \gets \textbf{BelUpdate}(a_t,o_t)$
   \UNTIL{$Episode Ends$}
\end{algorithmic}
\end{algorithm}

The core-test projection algorithm does not work well. The way we choose $\Upsilon_i(b^T)$ as $\sum b_p^{S_i|T}$ is flawed. Choosing a source task that has the core tests with maximum probability of occurrence in the target does not work well because core tests are not necessarily the most 'important tests' of a task. By 'important tests' we mean that higher probability of occurrence of these tests in the two tasks implies maximum similarity.

As a solution we introduce a set of tests called \textbf{validating tests}. These tests signify, that equal probability of occurrence of these tests, in source and target is geared towards optimal transfer.

\subsubsection{Validating Test Projection Algorithm}

Instead of projecting the target onto the core-test of source, we now project both the source and the target onto a common set of validating tests. We then find the distance between their projections and draw a conclusion on their similarity. 

Remember that from the PSR model that the probability of any test occurring given the current belief is $P(t|h) = P(Q|h)m_{t}$. Let the validating test be $V_t = \{t_{v_1},t_{v_2},t_{v_3}...,t_{v_m}\}$. Hence the projection of a task $T$ onto these tests is given by $P^T(V_t|h) = M^T_{V_t}P^T(Q|h)$. Value of $m^k_t$ for any test $t$ and any task $k$ is calculated using Eq.\ref{mo}. Let $\chi^{S}(V_t,h)$ be the projection of task $S$ onto validating tests $V_t$ after observing history $h$. We define the distance between source($S$) and target($T$) as the dot product between their projections on validation tests given by, $$\Upsilon_s(b^s(h^s),b^T(h^T)) \coloneqq \chi^S(V^t,h^s)\cdot\chi^T(V_t,h^T)$$

Even for two similar tasks, their projections depend on their histories. We assume that there exists a history in every source task which when applied results in a similar configuration as in target task. Applying a history to a source means updating the initial belief of the source using the history as observed trajectory. We call the history that maximizes the similarity between a source $s$ and the target while at the same time has a high probability of occurrence as \textbf{history offset} $(H^{s}_o)$ of that source. Mathematically,

\begin{equation}
H^s_{o}= \arg \max_{h^s} \frac{m_{h^s}.b^s(\emptyset)}{\chi^S(V_t,h^s)\cdot\chi^T(V_t,h)}
\label{s1}
\end{equation}

\begin{algorithm}
    \caption{Validating Test Projection Algorithm}
   \label{proj2}
\begin{algorithmic}
   \STATE {\bfseries Input:} $M^T_{V_t},M^S_{V_t},M^S,b^S(\emptyset),b^T(h),V_t$
   \STATE $\chi^T(V_t,h) \gets M^T_{V_t}.b^T(h)$
   \FOR{$b^{S_i}(\emptyset)$  in $b^S(\emptyset)$}
   \STATE $H^{i}_{o} \gets \textbf{HistoryOffset}(b^{S_i}(\emptyset))$ \COMMENT{Genetic Algo.}
   \STATE $b^{S_i}(H^{i}_{o}) \gets \textbf{BelUpdate}(b^{S_i}(\emptyset),H^{i}_{o})$
   \STATE $\chi^{S_i}(V_t,H^{i}_{o}) \gets M^{S_i}_{V_t}.b^{S_i}(H^i_{o})$
   \STATE $\mathcal{D}({S_i}) \gets \textbf{similarity}(\chi^{S_i}(V_t,H^{i}_{o}),\chi^T(V_t,h))$ \COMMENT{\ref{s2}}
   \ENDFOR
   \STATE $S^* \gets \arg \max_{S_i} \mathcal{D}$
   \STATE $action \gets \arg \max_a \mathcal{Q}(b^{S^*}(H^{S^*}_{o}),a)$
 \STATE \textbf{return} action
\end{algorithmic}
\end{algorithm}

The numerator of Eq.\ref{s1} is the probability of history occurring given the initial state, $P^s(h|\emptyset)$. The denominator is the distance between the source and target as we discussed in the previous paragraph. A sub-problem of finding similarity would be to find the history offset of every source task with respect to the target. Solving Eq.\ref{s1} is formulated as a search problem. In this problem the search space is huge. For a history of length $n$ the search space is exponential in action observation space $O(\mid \mathcal{A} \times \mathcal{O}\mid^n)$. We use genetic algorithm to search through this huge space as the solution is confined to a very small part the space and there are potentially many such solutions i.e. more than one history can result in same belief state.

For a given task, the algorithm generates random trajectories of random lengths. The length of each trajectory is limited to the rank of system dynamics matrix of that task. It generates $\mathcal{P}$ such trajectories and initialize them as initial population. The genetic algorithm uses the function under maximization, as the fitness function Eq.\ref{s2}.
\begin{equation}
\Upsilon(b^s(h^s),b^T) = \frac{m_{h^s}.b^s(\emptyset)}{\chi^s(V_t,h^s)\cdot\chi^T(V_t,h)}
\label{s2}
\end{equation}

Since we find the history offset at every time step in target, we use current belief for the target in \ref{s2} that is obtained after observing the current history $h$ in target.

The genetic algorithm crosses-over between two selected individuals from the population based on their fitness. Since the fitness of any individual lies between 0 and 1, $0 \le f(s) \le 1$, we cannot get high variance in the weights (for prominent selection of individuals with high fitness when weighted sampling is used). Hence the fitness of every individual is reassigned to the rank of every individual (ranked based on their current fitness). We select (weighted selection using our updated fitness) $\mathcal{P}$ pairs or $2\mathcal{P}$ individuals for crossover. The cut-point $c$ is chosen between 1 through maximum length of two parents under crossover using weighted-random sampling. Encouraging histories of shorter lengths, the algorithm gives more weights to 1 through minimum length of the parents. Crossing over at the cut point gives $2\mathcal{P}$ children. With a fixed probability our method tweaks the actions and observations in the children. Also with a small probability it adds selected $ao$ (one time step action-observation trajectory) at the end of an individual. This one-step $ao$ is either randomly selected or copied from the last action-observation of an individual. For rejection, the algorithm rejects $\mathcal{P}$ of $2\mathcal{P}$ such individuals based on their fitness. Next, it reassigns the current population to the selected children. We repeat this algorithm for a fixed number of generations. The individual with maximum fitness is taken as the history offset for the source under search.

At every time-step our algorithm finds the history offset $H^i_{o}$ of every source $S_i$. We apply this history offset to the initial belief of the corresponding sources. Mathematically,  $b^{S_i}(\emptyset) \xrightarrow{H^i_{o}} b^{S_i} (H^i_{o})$. The source to transfer policy from is selected using Eq.\ref{s2} as $$ S^* = \arg \max_s \Upsilon(b^s(H^s_{o}),b^T)$$ Finally,  the policy from $S^*$ after applying its history offset $H^{S^*}_{o}$ is transferred at this time-step. $$\pi^*(b^T(h)) = \arg \max_a \mathcal{Q}(b^{S^*}(H^{S^*}_{o}),a) $$

Thus, we approximate $b^{*S|T}$ as $b^{S^*}(\mathcal{H}^{S^*}_o)$

\section{EXPERIMENTS AND RESULTS}

\begin{figure}[H]
\begin{subfigure}
\centering
\includegraphics[width=0.5\textwidth]{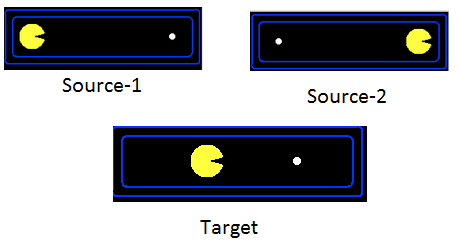}
\caption{Test domain 1 (SET-1)}
\label{set1}
\end{subfigure}

\begin{subfigure}
\centering
\includegraphics[width=0.5\textwidth]{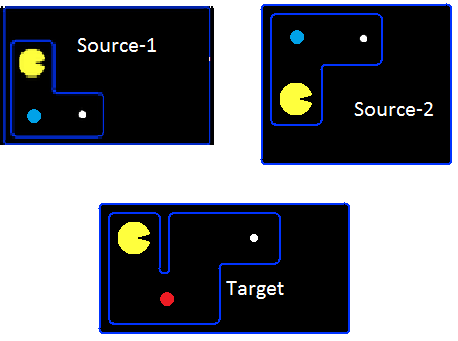}
\caption{Test domain 2 (SET-2).  \textbf{Note:} Red and Blue dots are used only for marking. They are not pellets or ghosts. The pellets are shown as white dots.}
\label{set2}
\end{subfigure}
\end{figure}

\paragraph{}
We evaluated our proposed transfer algorithm on two different sets (SET-1 and SET-2) of partially observable mazes depicted in Fig.\ref{set1} \& \ref{set2}. The target task of SET-1 and SET-2 have 6 and 8 free cells respectively. We use partially observable Pacman (Pocman) with one pellet and no ghosts (we consider only navigation through a maze). The agent cannot sense its exact position but can sense the presence of wall in specified directions. We simplified the environment and dimensions in observation space for the ADL to model the world in reasonable time. In SET-1 we sense in west and east directions while in SET-2 we sense in north and west directions. Hence the size of observation space is 4. Every sensor (in each direction) reads the correct value with probability 0.95. At each time step, the agent takes an action in one of the four directions. The action gets executed with probability 0.9 otherwise the agent moves in one of the perpendicular directions. If the agent bumps into a wall it remains in place. The agent receives a -1 reward at every time-step except at goal state where it receives +10. 

For a given set of tasks, our algorithm learns the PSR model of each task from their POMDP model using ADL. We generated 100 trajectories each of length 50 in every source task. The trajectories are used to train CMAC for approximate Q-learning. The $\mathcal{Q}$-learning algorithm is run using a completely random policy. The action edge of the CMAC is discretized into $\mid \mathcal{A} \mid$ equal segments, 4 here. Every component of the prediction vector is the probability of core test occurring given history $P(Q \mid h)$ and ranges between 0 and 1. In our case (SET-2) the source task and target task have 4 and 8 core tests respectively. For example, the edge representing the prediction vector component of CMAC for the target task is quantized in steps of 0.125. Also there are 8 overlapping hypercube tilings offset by 0.015625 units. The tile values initialized randomly and the update is applied online.

In genetic algorithm, the maximum size of every individual in the population limited is limited to 10. The best individuals for crossover was selected with a probability 0.8. The probability of mutation was fixed to 0.15. During mutation we either randomly tweak the action, observations in individuals or insert or delete elements from individuals with probability 0.5. Finally, individuals with low fitness are rejected with a probability of 0.9. At every iteration our algorithm deleted randomly selected individuals and initialized them to random histories to keep the diversity up in population. 

In SET-2, we used $\{[(S2-1)^2(E1-1)^2], [(N2-1)(N0-1)(E1-1)(E1-1)] , [(E1-1)^2] ,[(N2-1)(N0-1)(E2-1)(E2+10)]\}$ as the validating set of tests. There are four tests in the set. The first element in each test is an action followed by observation and reward. For example $(S2-1)$ denotes taking the action 'South' and observing 2 (binary equivalent: '10' meaning we observe no wall in North and a wall in West) and a reward -1. Given the initial states as shown in Fig.\ref{set2}, the genetic algorithm returns history offsets of Source-1 and Source-2 as $(N0-1))$ and $(E2-1)$ respectively.

\begin{figure}[h]
\begin{tikzpicture}
\begin{axis}[
  xlabel=State,
  ylabel=Similarity of target state with source $\Upsilon_i(b^T)$,
  xtick={1,2,3,4,5,6,7,8},
  legend pos=south east]
\addplot table [y=$A$, x=S]{simi.dat};
\addlegendentry{Source 1}
\addplot table [y=$B$, x=S]{simi.dat};
\addlegendentry{Source 2}
\end{axis}
\end{tikzpicture}
\caption{Graph showing the similarity of source to target.}
\label{g1}
\end{figure}
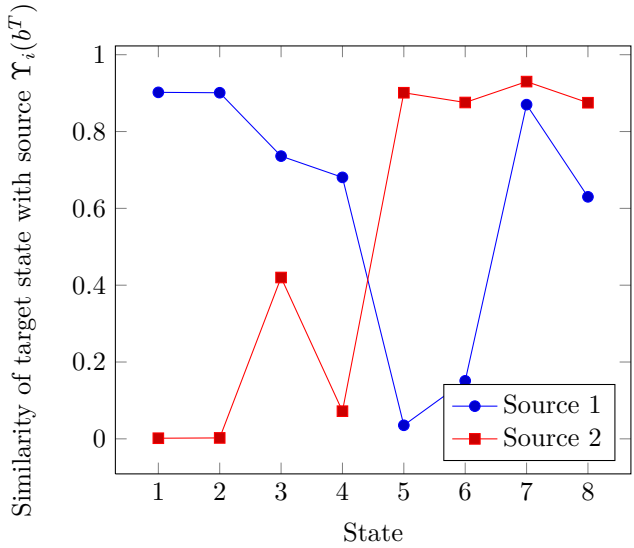

After this update on initial beliefs of both the sources with their corresponding history offsets and finding their respective projections on the set of validating tests, the projection of $Source-1$ was close to the projection of target. In fact, the correlation between the projections of $Source-1$ and target was 0.902 while that of $Source-2$ and target was 0.001. Given the updated (after applying history offset) belief of $Source-1$, the Q-function returned the action 'South', which is the expected optimal action given the initial state in target Fig.\ref{set2} (SET-2). When the position of Pocman was at the red dot (Fig.\ref{set2}), the history offsets of Source-1 and Source-2 were $(S2-1)(S2-1)$ and $(N2-1)(W2-1)(N0-1)$. After applying these offsets to their respective sources (the Pocman will be at the blue dots. Fig.\ref{set2}), we see that the optimal action in both these environment is 'East'. Which implies we must observe almost equal similarity between both the sources and target. In fact, the similarity of $Source-1$ and $Source-2$ were 0.73 and 0.42 respectively. The results of similarity vs different states is plotted in Fig.$\ref{g1}$.

\begin{figure}
\centering
\includegraphics{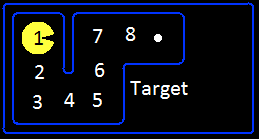}
\caption{Numbering for Pocman states}
\end{figure}

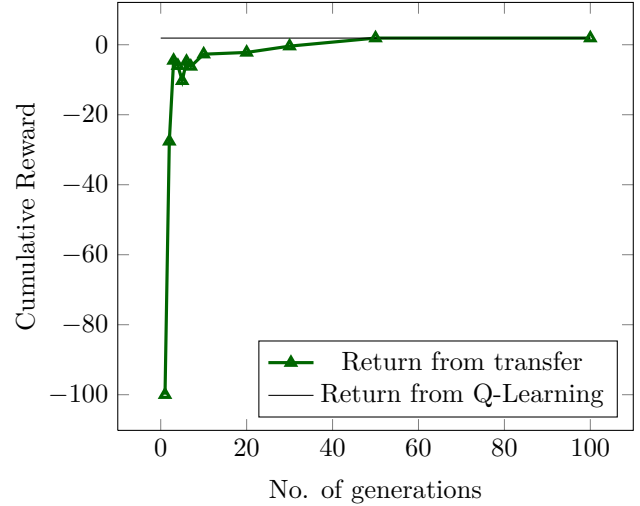
\begin{figure}[h]
\begin{tikzpicture}
\begin{axis}[
  xlabel=No. of generations,
  ylabel=Cumulative Reward, legend pos=south east ]
\addplot[color=green!40!black, very thick, mark=triangle] table [y=$A$, x=I]{popcons.dat};
\addlegendentry{Return from transfer}
\addplot[mark=none, black] coordinates {(0,1.9) (100,1.9)};
\addlegendentry{Return from Q-Learning}
\end{axis}
\end{tikzpicture}
\caption{GA for constant population size, (50).}
\label{g2}
\end{figure}

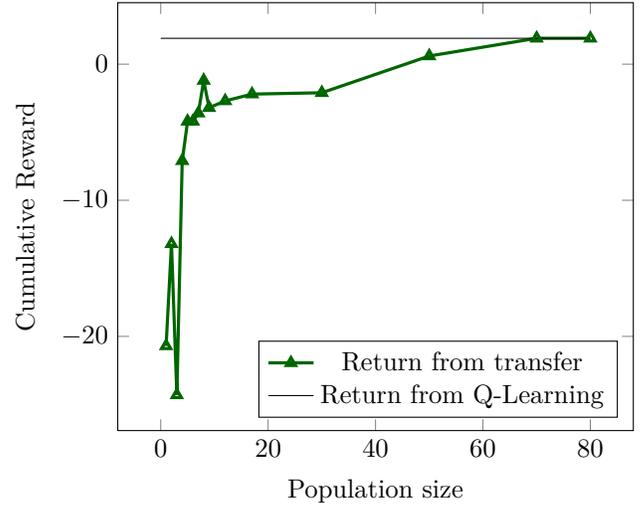
\begin{figure}[h]
\begin{tikzpicture}
\begin{axis}[
  xlabel=Population size,
  ylabel=Cumulative Reward, legend pos=south east ]
\addplot[color=green!40!black, very thick, mark=triangle] table [y=$A$, x=P]{itercons.dat};
\addlegendentry{Return from transfer}
\addplot[mark=none, black] coordinates {(0,1.9) (80,1.9)};
\addlegendentry{Return from Q-Learning}
\end{axis}
\end{tikzpicture}
\caption{GA for constant no. of generations, (30).}
\label{g3}
\end{figure}

To illustrate the validity of our approach we compare the cumulative reward received by following policy returned from planning in target environment (regular approach) and that returned from our transfer algorithm. The benchmark for our transfer is to converge to the cumulative reward received from planning. The cumulative reward (averaged over 10 trials) against the no. of generations and population is given in Fig \ref{g2}. Fig. \ref{g2} was plotted with constant population, $50$ and Fig. \ref{g3} was plotted with constant generations, $30$. We ran the algorithm till 100 time-steps or till the Pocman emerges victorious, whichever happened first. Hence the maximum possible return is +2 while the minimum possible return is -100. Considering the limited time available for genetic algorithm at every time-step, we used generations = 30, population = 50 throughout our experiments.

\section{FUTURE WORK AND DISCUSSIONS}

The results we obtained were promising. However more empirical results from other domains is needed. Using the ADL algorithm we were able to model only simple Pocman environments with small observation spaces. The same transfer algorithm should be tested using models of complex environment. TPSR [7] provides a good framework for modeling complex environments. Also we have assumed that the cost of learning the model to be zero or that the model is given to us. Our next step would be to improve our algorithm for online learning where the model of the target task is not known beforehand. Interleaved learning and planning [6] provides a method to simultaneously learn and plan using PSRs. One important direction of research is to incorporate the transfer algorithm into this framework. 

We used hand-coded validation tests for SET-1 and SET-2. We are currently investigating ways to automatically find a good set of validating tests given the source tasks. One simple example to show the limitations of our approach would be a source Pocman where the optimal action consist of only "East", and a target Pocman where the optimal action consists of only "North". Our current approach does not address the problem of finding the appropriate action mappings. We are working an approach to map a given action-observation in target to an action-observation in source. At every time-step we can find the change in history offset and relate this change to action-observation in target. This shall provide the required mapping. With this mapping finding the history offset at every time step is unnecessary.

We also plan on searching in beliefs space instead of histories, since this would lead to a smaller search spaces or we could even find a closed form solution for the optimal belief. One future work could be to extend the bisimulation metrics to PSRs and use it to transfer the policy.

\section{CONCLUSION}

This is the first work for transfer in partially observable environments. We develop a basic framework for the problem. We show in simple experiments the successful transfer of policy to solve a partially observable maze. Our method was able to find similarity (quantitatively) between tasks and was able to transfer policy from the most similar source task. 

Our method is suitable when there is a reachable configuration in one of the source tasks that is similar to the target task and has the same optimal action (as in target task). In case if all sources have no such configuration or no common optimal action, the algorithm needs an action map between the sources and the target.

\subsubsection*{Acknowledgements}

We thank Prof. Dmitry Berenson for his constant motivation and constructive feedback throughout the project. We also extend our gratitude to Prof. Balaraman Ravindran, Mr. Prasanna Parthasarathi and Mr. Janarthanan Rajendran from IIT Madras.

% \bibliography{example_paper}
% \bibliographystyle{abbrvnat.bst}
%\bibliographystyle{natbib.sty}

% Use unnumbered third level headings for the acknowledgements title.
% All acknowledgements go at the end of the paper.

\subsubsection*{References}

% % References follow the acknowledgements.  Use unnumbered third level
% % heading for the references title.  Any choice of citation style is
% % acceptable as long as you are consistent.

\begin{enumerate}
    \item J.~S.~Albus.  A theory of cerebellar function. \textit{Mathematical Biosciences}, 10(1):25-61, 1971.
    
    \item B.~Boots,  S.~M.~Siddiqi,  and  G.~J.~Gordon.   Closing the learning-planning loop with predictive state representations. \textit{The International Journal of Robotics Research}, 30(7):954-966, 2011.
    
    \item P.~S.~Castro. On planning, prediction and knowledge transfer in fully and partially observable markov decision processes.\textit{ McGill University,  PhD thesis,} 2011.
    
    \item M.~R.~James and S.~Singh. Learning and discovery of predictive state representations in dynamical systems with reset. In \textit{Proceedings of the  twenty-first international conference on Machine learning}, page 53. ACM, 2004.
    
    \item M.~R.~James and S.~Singh. Planning with predictive state representations. \textit{IEEE  Proceedings. 2004 International Conference on Machine Learning and Applications}, pages 304-311, 2004.
    
    \item S. Ong, Y. Grinberg, and J. Pineau. Goal-directed online learning of predictive models.  In \textit{Recent Advances in Reinforcement Learning, Volume 7188 of Lecture Notes in Computer Science},  Pages 18-29. Springer Berlin Heidelberg, 2012. ISBN 978-3-642-29945-2.
    
    \item M.~Rosencrantz, G. Gordon, and S. Thrun. Learning low dimensional predictive representations. In \textit{Proceedings of the twenty-first international conference on Machine learning}, page 88. ACM, 2004.
    
    \item M.~E.~Taylor and P.~Stone. An introduction to inter-task transfer for reinforcement learning. \textit{AI Magazine}, 32(1):15-34, 2011.
    
    \item C.~J.~C.~H.~Watkins. Learning from delayed rewards. \textit{PhD thesis, University of Cambridge, England}, 1989.
    
    \item B.~D.~Wolfe. Modeling dynamical systems with structured predictive state representations. \textit{PhD thesis, University of Michigan}, 2009.

    \item M.~L.~Littman, R.~S.~Sutton, S.~Singh. Predictive representations of state. Advances in Neural Information Processing Systems 14 (NIPS), pages  1555-1561,2002    
\end{enumerate}

% J.~Alspector, B.~Gupta, and R.~B.~Allen  (1989). Performance of a
% stochastic learning microchip.  In D. S. Touretzky (ed.), {\it Advances
% in Neural Information Processing Systems 1}, 748-760.  San Mateo, Calif.:
% Morgan Kaufmann.

% F.~Rosenblatt (1962). {\it Principles of Neurodynamics.} Washington,
% D.C.: Spartan Books.

% G.~Tesauro (1989). Neurogammon wins computer Olympiad.  {\it Neural
% Computation} {\bf 1}(3):321-323.

\end{document}